# Dynamic Bayesian Multinets

Jeff A. Bilmes
Department of Electrical Engineering
University of Washington
Seattle, WA 98195-2500

## Abstract

In this work, dynamic Bayesian multinets are introduced where a Markov chain state at time $t$ determines conditional independence patterns between random variables lying within a local time window surrounding $t$. It is shown how information-theoretic criterion functions can be used to induce sparse, discriminative, and class-conditional network structures that yield an optimal approximation to the class posterior probability, and therefore are useful for the classification task. Using a new structure learning heuristic, the resulting models are tested on a medium-vocabulary isolated-word speech recognition task. It is demonstrated that these discriminatively structured dynamic Bayesian multinets, when trained in a maximum likelihood setting using EM, can outperform both HMMs and other dynamic Bayesian networks with a similar number of parameters.

## 1 Introduction

While Markov chains are sometimes a useful model for sequences, such simple independence assumptions can lead to poor representations of real processes. An alternative and highly successful extension to the Markov chain allows random functions to be applied to each Markov state to yield the hidden Markov model (HMM). As is well known, an HMM is simply one type of dynamic Bayesian network (DBN) [12], or more generally a graphical model [21]. When HMMs are considered as one small instance in this enormous family of models, it is reasonable to suppose that the independence assumptions underlying an HMM can further be relaxed to yield better models still.

The structure of a graphical model, however, is sometimes chosen for an application without ensuring that it matches the underlying process the model is supposed to represent. Using a hidden Markov model to represent speech, for example, is one such instance of pre-specifying an arbitrary model structure for a domain. Such an approach has obvious computational and infrastructural advantages: if the model is kept simple, inference is guaranteed to stay tractable and software tools can be developed and reused many times. The conditional independence properties of a particular model, however, could be sub-optimal for a given task. With a more appropriate model, substantial improvements in classification accuracy, memory requirements, and computational demands could potentially be achieved. While it might be sufficient to hand-specify the model for a given application, a promising approach allows the data itself to determine or at least influence the model.

In the most general case, there are four distinct components of a graphical model: the *semantics*, the *structure*, the *implementation*, and the *parameters*. There are a variety of different semantics, including directed (Bayesian network) and undirected (Markov random field) models, chain graphs, and other more experimental frameworks. In general, each corresponds to a different family of probability distributions and, based on training data, a semantics could potentially be selected or perhaps even induced anew. Fixing the semantics, obtaining a good model structure is crucial, and is therefore a current active research focus [7, 16, 5, 20, 10]. Fixing the structure, there are a variety of ways to implement[1] the dependencies between random variables, such as conditional probability tables, neural networks, decision trees, Gaussian polynomial regression, and so on. And finally, fixing all of the above, a good assignment of all the parameters must be found. Of course in each case, a Bayesian approach can also be taken where we use a (potentially uncountably infinite) probabilistically weighted mixture over multiple choices.

The task of learning graphical models can be seen as learning any or all of the above four components given a collection of data, and is akin to model selection [22] problem known to the statistics community for years. In all cases, the underlying goal is to identify a system for probabilistic inference that is computationally efficient, accurate, and somehow informative about the given problem domain.

---

[1] While this is not standard terminology, a concise way to refer to the representation of the local conditional probability model is simply to use the term "implementation."

In this paper, a class of models called dynamic Bayesian multinets and a method to induce their structure for the classification task is described. In this work, an extension of [3], the problem domain is speech recognition so it is necessary to use dynamic models. Also, since classification is the goal, it is beneficial to learn class-specific and (as we will see) discriminative structure. And to further improve sparsity (and therefore reduce computational and memory demands) and to represent class conditional information only where necessary, Bayesian multinets (described in the next section) are used.

Section 2 provides a review of structure learning in Bayesian networks, of Bayesian multinets, and presents the idea of structural discriminability. Section 3 introduces the class of models considered in this work and analyzes their inferential complexity. Section 4 provides three information theoretic criterion functions that can be used to learn structure, the last of which provably provides an optimal approximation to the local posterior probability. Section 5 introduces the improved pairwise algorithm, a heuristic developed because the above induction procedure is computationally infeasible. Section 6 evaluates this system on a medium-vocabulary speech corpus and shows that when structure is determined using the discriminative induction method and trained using EM, these networks can outperform both HMMs and other dynamic Bayesian networks with a similar number of parameters. But when structure is determined arbitrarily, or without using a discriminative method, the performance is dramatically worse. Finally, Section 7 concludes and discusses future work.

## 2 Background

### 2.1 Structure Learning

A fully-connected graphical model can represent any probability distribution representable by a sparsely structured one, but there are many important reasons for not using such a fully connected model. These include 1) sparse network structures have fewer computational and memory requirements; 2) a sparse network is less susceptible to noise in training data (i.e., lower variance) and less prone to overfitting; and 3) the resulting structure might reveal high-level knowledge about the underlying problem domain that was previously drowned out by many extra dependencies. A graphical model should represent a dependence between two random variables only when necessary, where "necessary" depends on the current task. In essence, learning the structure in data is similar to developing an efficient code for the underlying random process, as efficient coding is analogous to probabilistic modeling.

Perhaps the earliest well-known work on structure learning in directed graphical models is [7]. More recent research on this topic may be found in [17, 5, 16, 25, 10, 20, 23, 13].[2] In general, the task of learning Bayesian networks can be grouped into four categories [10] depending on 1) if the data is fully observable or if it contains missing values, and 2) if it is assumed that the structure of the model is known or not. The easiest case is when the data is fully observable and model structure is known, whereas the most difficult case is when the data is only partially observable and when the structure is unknown or only partially known.

Note that a general optimization procedure can be used to learn many aspects of a graphical model. Often, learning needs only a maximum likelihood procedure perhaps with an additional complexity penalty term such as MDL or BIC. Alternatively, a Bayesian approach to learning can be used where no single structure or set of parameters are chosen. For certain classes of networks, the prior and posterior are particularly simple [16]. Alternatively, a risk minimization approach [26] can be applied to the learning problem.

In principle, an optimization procedure could simultaneously cover all four components of a graphical model: semantics, structure, implementation, and parameters. There has, however, been little if any research on methods to learn the best implementation and semantics. The problem becomes inherently difficult because the quality of each component cannot be accurately evaluated without first obtaining good settings for the other three components. The problem becomes more arduous when one begins to consider multi-implementation and multi-semantic models. In practice, therefore, one or more components are typically fixed before any optimization begins.

### 2.2 Bayesian Multinets

A advantage of Bayesian networks is that they can specify dependencies only when necessary, leading to a significant reduction in the cost of inference. Bayesian multinets [15, 14] further generalize Bayesian networks and can further reduce computation. A multinet can be thought of as a network where edges can appear or disappear depending on the values of certain nodes in the graph, a notion that has been called asymmetric independence assertions [14].

Consider a network with four nodes $A$, $B$, $C$ and $Q$. In a multinet, the conditional independence properties among $A$, $B$, and $C$ might, for example, change for differing values of $Q$. If $Q$ is binary, and $C \perp\!\!\!\perp A | \{B, Q = 0\}$ but $C \not\!\perp\!\!\!\perp A | \{B, Q = 1\}$, then the joint probability could be writ-

---

[2]See, especially, the reviews given in [16, 5, 20].

ten as:

$$p(A, B, C) = \sum_q p(A, B, C|Q = q)p(Q = q)$$
$$= p(C|B, Q = 0)p(B|A, Q = 0)p(Q = 0) +$$
$$p(C|B, A, Q = 1)p(B|A, Q = 1)p(Q = 1)$$

Some examples of multinets include mixtures of tree-dependent distributions [23] and class-conditional naive Bayes classifiers [11, 10].

In general, the statistical dependencies in a multinet could be represented by a regular Bayesian network via specific values of the parameters [14] (e.g., for switching linear Gaussian models, certain parameters could be zero, or for discrete probability tables, hyperplanes could indicate independence between random variables only for certain values of other random variables). In other words, the family of probability distributions representable by Bayesian networks and by Bayesian multinets is the same. In practice, however, a multinet could result in a substantial savings in memory, computation, and necessary sample-size complexity relative to an equivalent Bayesian network.

## 2.3 The Classification Task

Many papers on structure learning concentrate on producing networks that best represent statistical dependencies extant in data. When the goal is classification, however, this is not necessarily optimal. Indeed, the class posterior probability will be accurately approximated if sample and class label are considered together, and then jointly optimized in a maximum likelihood procedure, assuming sufficient data.

Such a procedure might be wasteful, however, as likelihood scores are penalized from the term containing dependencies only between features (which has a much larger magnitude) than the term containing the class posterior probability. It is this later term that, according to Bayes decision theory, must be accurately modeled to achieve good classification performance. In [10], this issue was noticed, and both extended versions of naive Bayes classifiers and class conditional Bayesian multinets were considered, both of which outperformed the naive Bayes classifier on classification tasks.

In a general classification task, additional reductions in computation and increases in sparsity can be achieved by learning a specific network structure for each class, where each class-conditional network represents nothing other than those dependencies, often unique to its class, that help approximate the class posterior probability. This property has been called *structural discriminability* [2].

## 3 The Model

In this work, we consider a class of models called dynamic Bayesian multinets (DBM). They consist of hidden Markov chains that determine local class-conditional Bayesian networks over a window of observations. Equivalently, they consist of extensions to hidden Markov models (HMMs) where additional cross-observation dependencies have been added as a function of the underlying Markov chain. This model is also called a buried Markov model (BMM) [3] because the hidden Markov chain in a HMM is further hidden (buried) by additional cross-observation dependencies.

First some notation: $Q_t$ refers to a Markov state at time $t$, and $Q_{1:T} \triangleq \{Q_1, Q_2, \ldots, Q_T\}$ refers to the entire chain.[3] $X_t$ will refer to the observation vector at time $t$ with $X_{ti}$ referring to its $i^{th}$ element. Using this notation, a hidden Markov model is a collection of hidden $Q_{1:T}$ and observation $X_{1:T}$ variables that possess the following conditional independence properties: $\{X_{t:T}, Q_{t:T}\} \perp\!\!\!\perp \{Q_{1:t-2}, X_{1:t-1}\}|Q_{t-1}$ and $X_t \perp\!\!\!\perp \{Q_{\neg t}, X_{\neg t}\}|Q_t$ for all $t$.

We generalize this model such that $X_{ti}$ is no longer conditionally independent of all the surrounding observations given $Q_t$. Relative to an HMM, a DBM has been augmented with chain-conditional cross-observation dependencies between individual observation elements. The probability model becomes:

$$p(x_{1:t}) = \sum_{q_{1:t}} \prod_t p(x_t|z_t(q_t), q_t)p(q_t|q_{t-1})$$

where $z_t(q) \subseteq x_{<t}$ for all $t$ and $q$. For example, it could be that $z_t(q) = \{x_{t-1,3}, x_{t-1,5}, x_{t-2,1}, x_{t-3,9}\}$ and $z_t(r) = \{x_{t-4,2}, x_{t-9,0}\}$ for $r \neq q$. A multinet occurs because $z_t(q)$ is a function of $q$; if the Markov chain changes, so will the set of dependencies. Specifically, the conditional independence assumption among observation elements becomes $X_{ti} \perp\!\!\!\perp \{X_{\neg t} \setminus z_{ti}(q)\}|\{q, z_{ti}(q)\}$. This class of model is depicted in Figure 1 for two instantiations of the Markov chain.

In general, adding conditional dependencies in a DBN can significantly increase computational and memory complexity. For the junction tree algorithm, the complexity is $O(\sum_{i=1}^T s(C_i))$ where $T$ is the number of resultant cliques in the junction tree, and $s(C_i)$ is the size of the state space for clique $C_i$. For an HMM with $T$ time-steps and $N$ states, there are $O(T)$ cliques each with at most a state space size of $N^2$ resulting in $O(TN^2)$.

To determine the complexity of a DBM, first define [4] an

---

[3] In general, $X_{1:t}$ is matlab-like notation to refer to the set of variables with indices between 1 and $t$ inclusive, $X_{<t} \triangleq X_{1:(t-1)}$, and $X_{\neg t} \triangleq \{X_{1:T} \setminus X_t\}$.

[4] AR-HMM stands for auto-regressive HMM. An AR-HMM

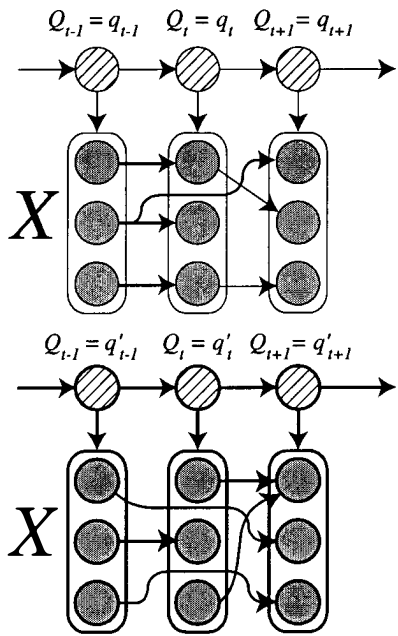

Figure 1: Two networks corresponding to two values of the Markov chain, $q_{1:T}$ and $q'_{1:T}$. The individual elements of the observation vectors are shown explicitly as dark shaded nodes surrounded by boxes. Nodes with diagonal lines correspond to instantiated hidden variables.

AR-HMM($K$) as an HMM but with additional edges pointing from observations $X_{t-\ell}$ to $X_t$ for $\ell = 1, \ldots, k$ (See Figure 2). Collapsing the vector graphical notation in Figure 1 into a single node, and including a dependency from observations $X_\tau$ to $X_t$ in the AR-HMM if for any value of $Q_t$ and there exists a dependency between $X_{\tau i}$ and $X_{tj}$ for any $i, j$, a DBM with a maximal dependency across $K$ observations can be more generally represented by an AR-HMM($K$).

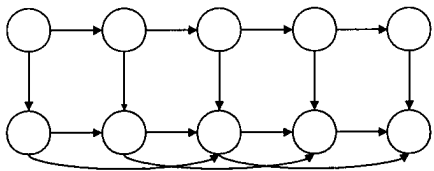

Figure 2: Bayesian network for an AR-HMM(2)

There are three things to note about an AR-HMM($K$). First, a moralized AR-HMM($K$) is triangulated. This can be seen by induction, the base case being obvious, and the induction step following because 1) a cycle containing edges only contained in the previous step's graph must have a chord by induction, and 2) a cycle containing edges not in the previous step's graph (new edges) must also have a chord because the portion of a cycle not containing

---

can of course use any, possibly non-linear, implementation between observations.

the new edges must go through a clique containing nodes adjacent by these new edges. Second, a triangulated-by-moralization AR-HMM($K$) has at most two hidden variables in its cliques since no node has more than one hidden variable as a parent, so moralization does not add edges between hidden variables. The remaining clique variables are observations, so the state-space size is only $N^2$. Third, such a triangulated AR-HMM($K$) has only $O(T)$ cliques. Therefore, the complexity of an AR-HMM($K$) and therefore a DBM is again only $O(TN^2)$ for any fixed $K$. There is, however, a constant cost associated with the number of additional dependency edges. Incorporating this cost, the complexity becomes $O(TN^2K)$ where $K$ is the maximum number of dependency edges per observation. The extra dependency structure is sparse, however, so the computational and memory requirements of a DBM will in practice be much less than its $O(TN^2K)$ complexity suggests.

## 4 Structure Learning in DBMs

Structure learning consists of optimally choosing $z_t(q)$ in $p(x_t|z_t(q), q)$. In this section, three methods are considered. In each case, a fixed upper bound is assumed on the possible number dependency variables. That is, it is assumed that $|z_t(q)| \leq c$ for some fixed $c > 0$. The phrases "choose dependency variable" and "choose dependencies" will be used synonymously. It is assumed that the reader is familiar with information-theoretic constructs [8].

The following theorem will be needed:

**Theorem 4.1. Mutual Information and Likelihood.** *Given three random variables $X$, $Z^{(a)}$ and $Z^{(b)}$, where $I(X; Z^{(a)}) > I(X; Z^{(b)})$, the likelihood of $X$ given $Z^{(a)}$ is higher than given $Z^{(b)}$, for $n$, the sample size, large enough, i.e.,*

$$\frac{1}{n}\sum_{i=1}^{n} \log p(x_i|z_i^{(a)}) > \frac{1}{n}\sum_{i=1}^{n} \log p(x_i|z_i^{(b)}) \quad (1)$$

*Proof.* Under the assumption, it immediately follows that:

$$H(X|Z^{(a)}) < H(X|Z^{(b)}).$$

Negating and expanding as integrals gives

$$\int \log p(x|z^{(a)}) dP(x, z^{(a)}) > \int \log p(x|z^{(b)}) dP(x, z^{(b)})$$

or equivalently

$$\lim_{n \to \infty} \frac{1}{n}\sum_{i=1}^{n} \log p(x_i|z_i^{(a)}) > \lim_{n \to \infty} \frac{1}{n}\sum_{i=1}^{n} \log p(x_i|z_i^{(b)})$$

where $(x_i, z_i^{(k)}) \sim p(X, Z^{(k)})$ for $k \in \{a, b\}$. The weak law of large numbers implies that $\forall \epsilon > 0$, $\exists n_a$ and $n_b$ such that for $n > \max(n_a, n_b)$,

$$\left| \frac{1}{n} \sum_{i=1}^{n} \log p(x_i | z_i^{(k)}) + H(X|Z^{(k)}) \right| < \epsilon$$

again for $k \in \{a, b\}$. Choosing $\epsilon < |H(X|Z^{(a)}) - H(X|Z^{(b)})|/2$ to get $n$ implies Equation 1. □

Of course, the actual probability distribution $p(x|z)$ is not known and only an approximation $\hat{p}(x|z, \Theta)$ is available where the parameters $\Theta$ are estimated, often using a method such as maximum likelihood, that decreases $D(p(x|z)||\hat{p}(x|z))$, the KL-distance between the actual and approximate distribution. If $\hat{p}$ is close enough to $p$, then the theorem above still holds. It is therefore assumed that the parameters of $\hat{p}$ have been estimated well enough so that any differences with $p$ are negligible.

The above theorem is, of course, also true for conditional mutual information [8] such as $I(X;Z|Q)$ or for a particular value of $q$, $I(X;Z|Q = q)$. Therefore, if $I(X;Z^{(a)}(q)|Q = q) > I(X;Z^{(b)}(q)|Q = q)$, for all $q$ then:

$$\frac{1}{T} \sum_{t=1}^{T} \log p(x_t | z_t^{(a)}(q_t), q_t) > \frac{1}{T} \sum_{t=1}^{T} \log p(x_t | z_{q_t}^{(b)}(q_t), q_t)$$

These quantities can be viewed as likelihoods of the data given Viterbi paths $q_t$ of modified HMMs. In the left case, the Viterbi path likelihood is higher. Note that using a similar argument as in the theorem, and because $H(X) \geq H(X|Z)$,

$$\frac{1}{T} \sum_{t=1}^{T} \log p(x_t | z_t^{(a)}(r_t), r_t) \geq \frac{1}{T} \sum_{t=1}^{T} \log p(x_t | r_t)$$

for some non-Viterbi path $r_t$ and for $n$ large enough. In other words, relative to an HMM, the likelihood of the data for paths other than the Viterbi path do not decrease when adding conditioning variables. The following theorem has therefore been shown.

**Theorem 4.2. A DBM with edges added relative to an HMM according to conditional mutual information produces a higher likelihood score than before modification.**

The DBM represents statistical relationships contained in the data that are not well represented before modification, which is the reason for the higher likelihood. Augmenting the dependencies according to conditional mutual information therefore defines the first dependency selection rule.

When the task is classification, however, a higher likelihood does not necessarily correspond to a lower error. Consider the two states $q$ and $r \neq q$. To achieve a lower error, a modification the $q$ and $r$ models should increase the average score of the $q$ model in the context of a sample from $q$ more than any increase in the $r$ model in the context of $q$ for all $r \neq q$. The score increases can in fact be negative, thereby decreasing the likelihood of both models, but potentially improving the classification accuracy. Accordingly, the score of a model in a different context can be evaluated using an extended form of conditional mutual information:[5]

$$I_r(X; Z|q) \triangleq \int p(x, z|r) \log \frac{p(x, z|q)}{p(x|q)p(z|q)} dx dz$$

Therefore, $I(X; Z(q)|q)$ should be large and $I_q(X; Z(r)|r)$ should not be as large for each $r$. This suggests optimizing the following:[6]

$$S(X; Z|Q) \triangleq \sum_{qr} p(q)(\delta_{qr} - p(r)) I_q(X; Z(r)|r)$$

where $Z = \cup_i Z(i)$ and $\delta_{qr}$ is a Kronecker delta. This quantity can be further motivated by noticing that the expected class posterior probability can be expanded as follows:

$$E[\log p(Q|X, Z)] = -H(Q|X, Z)$$
$$= -H(X|Q, Z) - H(Q|Z) + H(X|Z)$$

so that

$$E[\log p(Q|X, Z)] + H(X|Q) + H(Q) - H(X)$$
$$= I(X; Z|Q) + I(Q; Z) - I(X; Z)$$

Furthermore, the conditional entropy can be bounded by

$$-H(X|Z) = \int \log \left( \sum_r p(x|r, z) p(r|z) \right) dP(x, z, q)$$
$$\geq \int \sum_r p(r|z) \log p(x|r, z) dP(x, z, q)$$
$$\approx \sum_{r,q} p(r) p(q) \int \log p(x|z, r) dP(x, z|q)$$

where the first inequality follows from Jensen's inequality, and the approximate equality is valid if $I(Q; Z)$ is small. From this, it can be shown [2] that choosing $Z^{(a)}$ over $Z^{(b)}$ when $S(X; Z^{(a)}|Q) > S(X; Z^{(b)}|Q)$ will increase an upper bound on the expected class posterior probability, and therefore could potentially reduce the Bayes error. This defines a second dependency selection rule.

A generalization that does not require a small $I(Q; Z)$ can be obtained by noticing that $I(Q; Z)$ does not depend on $p(x|z, q)$. Therefore, if $Z = \cup_i Z(i)$ is chosen to maximize $I(X; Z|Q) - I(X; Z)$, the average class posterior probability $E[p(Q|X, Z)]$ will be maximized and therefore optimal for a fixed number of edges (see [2] for a proof). The quantity $I(X; Z|Q) - I(X; Z)$ could be called the *explaining away residual* (or the *EAR* measure), and it asks for edges that are more class-conditionally dependent than marginally independent. Moreover, a multinet can result when choosing class-conditional edges according to $I(X; Z|Q = q) - I(X; Z)$ for each $q$.

---
[5]Called cross-context conditional mutual information in [2].
[6]Called discriminative conditional mutual information in [2]

Summarizing this section, there are three possible rules that could be used to choose new edges between elements of $X_t$ and previous observations:

$$Z^*(q) = \underset{Z(q) \subseteq X_{<t} \;\&\; |Z(q)| \leq c}{\operatorname{argmax}} I(X; Z(q)|q) \quad (2)$$

$$Z^* = \underset{Z \subseteq X_{<t} \;\&\; |Z| \leq c}{\operatorname{argmax}} S(X; Z|Q) \quad (3)$$

$$Z^* = \underset{Z \subseteq X_{<t} \;\&\; |Z| \leq c}{\operatorname{argmax}} I(X; Z|Q) - I(X; Z) \quad (4)$$

Equations 3 and 4 produce discriminatively structured networks, in that the underlying dependencies represented by the network are unique to each class. The resulting models achieve a high score in the presence of a sample from the right class, but get a low score in the presence of a different class. More importantly, this can be true even for non-optimal parameter settings since, via the structure, the networks are inherently less capable of achieving high scores for samples of the wrong class. Therefore, along appropriate complexity penalties, it would be sufficient to learn parameters using likelihood based methods rather than the more costly risk-minimization procedures [26, 1, 9, 18, 19].

## 5 The Improved Pairwise Algorithm

The optimization suggested in the previous section is clearly impractical. In this section, a new computationally efficient heuristic, entitled the improved pairwise algorithm, is introduced. The algorithm approximates the desired quantities using only pairwise conditional mutual information between scalars.

The algorithm is presented in Figure 3 using rule 4. All candidate scalar random variables in $X_{1:t}$ are given and indexed by $Z_j$. A total of $M$ edges will be added separately for each value $q$, and for each $X_{ti}$. Therefore, the algorithm might allow for some redundancy if intra-feature dependencies are already modeled. The algorithm first sorts the scalars decreasing by the function $f(Z_j) = I(X_{ti}; Z_j|Q = q) - I(X_{ti}; Z_j)$. The output is $\mathbf{Z}_{qi}$, set of variables from which a link should be added to $X_{ti}$ under the class $q$.

The algorithm uses three criteria to eliminate candidate edges. The first ensures that the edge to $Z_j$ is actually informative about $X_{ti}$ in the context of $q$. The second is a redundancy check – it asks for an edge from a variable that has little information in common with the variables already added as depicted in Figure 4. The degree of allowed redundancy is determined using $0 < \tau < 1$. The third

---

**INPUT:** $Z_j, q, M, X_{ti}$
**OUTPUT:** $\mathbf{Z}_{qi}$
Set $\mathbf{Z}_{qi} = \emptyset$
Sort $Z_j$, so $f(Z_1) \geq f(Z_2) \geq \cdots$
For $j$ decreasing until $f(Z_j)$ falls below threshold:
    If $Z_j$ satisfies all the following three criteria:
        1) $I(X_{ti}; Z_j|q)$ is larger than a threshold
        2) For each $Z \in \mathbf{Z}_{qi}$, $I(Z_j; Z|Q) < \tau I(Z_j; X_{ti}|Q)$
        3) $I(X_{ti}; Z_j)$ is smaller than a threshold:
    then add $Z_j$ to $\mathbf{Z}_{qi}$ and break if $|\mathbf{Z}_{qi}| > M$.

Figure 3: The improved pairwise heuristic : this algorithm chooses the dependency variables for the $i^{th}$ feature position of $X_t$ and for class $q$.

and discriminative criterion ensures the candidate variable does improve the models scores in the general non-class-conditional case.

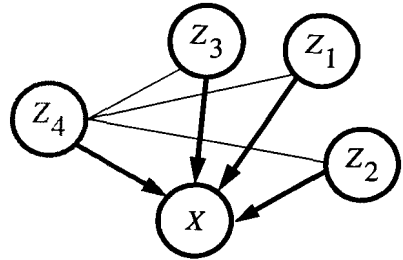

Figure 4: The redundancy check: The edge from $Z_4$ to $X$ is being considered. It will only be added if $I(Z_4; Z_i)$ for $i < 4$ is small, hopefully reducing the chance that a variable is added for which a class-conditional Markov blanket exists separating $Z_4$ from $X$ and making $Z_4$ superfluous.

This algorithm is inexpensive, running in time $O(|\mathcal{Q}||X|d^2)$ where $|\mathcal{Q}|$ is the number of classes, $|X|$ is the size of the observation vector, and $d \leq |X_{1:t}|$ is the maximum number of variables that are considered in the selection.

## 6 Experimental Evaluation

In this section, DBMs are evaluated in the context of classifying speech utterances. All experiments are reported using the NYNEX Phonebook database[24]. Phonebook is a large-vocabulary "phonetically-rich isolated-word telephone-speech database." It contains a rich collection of vocabulary words including poly-syllabic words such as "exhaustion," "immobilizing," "sluggishness," and "overambitious" as well monosyllabic words such as "awe," "biff," and "his."

The quantities $(X; Z|Q = q)$ are obtained using an initial baseline HMM-based system. The following general training procedure is used for all of the results reported in this

section:

1) Train a bootstrap HMM-based system using EM
2) Compute $(X; Z|Q = q)$ and $I(X; Z)$ for all pairs within a 200ms range surrounding the current time.
3) Run the improved pairwise heuristic
4) Train the resulting models again with EM
5) Test the result

The reported results all use a mixture of Gaussian linear-regression model to implement the dependencies. In this case,

$$p(x|z(q), q) = \left( \sum_{m=1}^{M} p(x|m, z(q), q) p(m|q) \right)$$

where each underlying component is a Gaussian linear-regressive process on $z$ using a sparse dependency matrix $B_{qm}$.

$$p(x|m, z, q) = \frac{1}{|2\pi \Sigma_{qm}|^{1/2}} e^{-\frac{1}{2}(x - B_{qm}z)' \Sigma_{qm}^{-1}(x - B_{qm}z)}$$

Therefore, $p(x|m, z, q)$ is a Gaussian distribution with conditional mean $B_{qm}z$ and covariance matrix $\Sigma_{qm}$. This implementation can, to some extent, simulate conditional variance by using mixtures, but it avoids many training complexities since closed-form EM update equations can be derived. Complete details of the experimental setup, training procedure, definitions of test and training sets, topology of the Markov chains, and so on are described in[2].

| |Vocab| | 75 | 150 | 300 | 600 | Params / SP |
|---|---|---|---|---|---|
| HMM | 5.0% | 7.0% | 9.2% | 13.3% | 157k / 1 |
| DBM | 4.6% | 6.5% | 8.7% | 12.3% | 157k / 1 |
| HMM | 2.4% | 3.5% | 5.6% | 7.7% | 182k /2 |
| DBM | 2.3% | 3.4% | 5.4% | 7.5% | 182k /2 |
| HMM | 1.7% | 2.8% | 4.6% | 6.2% | 163k /3 |
| DBM | 1.5% | 2.6% | 4.4% | 6.0% | 166k /3 |
| HMM | 1.5% | 2.7% | 4.3% | 5.8% | 200k /4 |
| DBM | 1.4% | 2.6% | 4.2% | 5.6% | 207k /4 |

Table 1: Word-error rate results comparing HMMs with DBMs. |Vocab| is the vocabulary size (number of test classes), Params is the number of system parameters, and SP is the number of Markov states per phoneme.

The results, given in Table 1, show the error rate for different topologies of Markov chain for each phone. The numbers are competitive with those reported using dynamic Bayesian networks on the same speech corpus and training/testing conditions [27], where best achieved result was 2.7% with a 515k parameter system on the 75-word vocabulary size case. The performance increases on average as more parameters are added to the system. In each case, however the DBM outperform an HMM with a comparable number of parameters and states per phone, and this is true across the different vocabulary sizes.

| Case | Type | WER | Params/ SP |
|---|---|---|---|
| 1 | CMI | 32.0% | 207k / 1 |
| 2 | AR2 | 27.6% | 207k / 1 |
| 3 | AR1 | 20.9% | 156k / 1 |
| 4 | RAND | 8.3% | 207k / 1 |

Table 2: Performance of DBMs determined using alternate methods and evaluated on the 75-word vocabulary test case.

A second set of results are given in Table 2 and can be compared to those given in the 75-word column in Table 1. Case 1 shows the performance of a DBM created using Equation 2. This rule adds dependencies that increase the model scores but not the classification accuracy. In fact, the likelihood scores in this case were dramatically larger than before modification. As can be seen, however, the performance dramatically decreases, presumably because the models are not structurally discriminative.

Cases 2 and 3 show the performance when dependencies are added from the previous (the two previous for case 3) observations to the current observation, and is therefore not a multinet. The performance is also very poor, indicating again that relaxing the wrong conditional independence properties can dramatically decrease classification accuracy.

Case 4 shows the performance when a different random set of dependencies between observation elements are added for each state. Interestingly, case 4 is much better than the previous cases suggesting that the most harmful and anti-discriminative dependencies have not been added. The performance, however, is still worse than the baseline HMM.

Several general points can be made from the two tables. Cases 1-3 indicate that dependencies that are added to a model structure to increase a (likelihood) score can cause a dramatic decrease in classification accuracy, even if the structures are augmented in a class-conditional way, as in case 1 above. Note that the likelihood scores for these models are dramatically higher both for the training *and testing* data, suggesting that overfitting is not the problem. The goal of many model selection methods [6, 22] is to choose a model that provides the best description of the data, but the above suggests that this can be inappropriate for classification. Admittedly, model selection procedures typically include complexity penalty terms (e.g., MDL, BIC, and so on). But these penalties do not select for discriminative structures.

Second, dependencies in a network should not be added just because they are missing. Cases 2 and 3 adds depen-dencies between adjacent observation vectors, an approach sometimes justified by noting that they are not directly represented by an HMM. But as the performance for these augmented models shows, the results indicate that adding missing dependencies can decrease classification performance.

Third, adding random dependencies does not produce as poor performance as in the previous cases, but neither is there any benefit. It is unlikely that choosing random dependencies, even if $q$-conditioned, will result in discriminative structure because the selection space is so large. The implications for structure learning methods that search over randomly chosen sets are clear: because of the large search space, it is unlikely that good sets of dependencies will be found in a reasonable amount of time. It seems crucial, therefore, to constrain the random search to those that found to be useful in some way, as has been argued in the past [13].

Finally, as argued in [2], an HMM can approximate a distribution arbitrarily well given enough capacity, enough training data, and enough computation. The results in the tables support this claim as increasing parameters leads to improved accuracy. The performance improvement obtained by adding more hidden states is dramatic, but the additional discriminative DBM dependencies can provide further improvements.

The results show that the DBM achieves the same or better classification performance with the same parameters, thereby supporting the claim that they have achieved sparser, higher performing, but lower complexity networks.

## 7 Discussion

In this paper, a class of graphical models is considered that generalizes HMMs. Several methods to automatically learn structure were presented that have optimal properties either by maximizing the likelihood score, or (the EAR measure) by maximizing the class posterior probability. A dependence selection heuristic, the improved pairwise algorithm, is introduced, and an implementation was tested using a medium-vocabulary speech corpus showing that appreciable gains can be obtained when the dependencies are chosen appropriately.

While this paper does not address the problem of learning the *hidden* structure in networks and uses only a simple Markov chain to represent dynamics [4, 12], for speech, it is often sufficient to consider a Markov chain as a probabilistic sequencer over strings of phonetic units. The multinets, which are conditioned on each of these sequences, determine local structure. Ultimately, it is planned to use and learn more complex models of dynamic behavior for those classes of signals that can benefit from it. It is also planned to use the EAR measure to determine more general discriminatively structured Bayesian multinets.

## Acknowledgements

This work has benefited from discussions with Nir Friedman, Steven Greenberg, Michael Jordan, Katrin Kirchhoff, Kevin Murphy, Stuart Russel, and Geoff Zweig. The author would like to acknowledge the International Computer Science Institute at which many of the computational experiments were performed, and thank the three anonymous reviewers for their useful comments.